\documentclass[a4paper,fleqn]{cas-dc}

\usepackage[authoryear]{natbib}

\usepackage{subfigure}
\usepackage{algorithm}
\usepackage{algorithmicx}
\usepackage{algpseudocode}
\usepackage{amsmath,amsfonts}
\usepackage{graphicx}

\def\tsc#1{\csdef{#1}{\textsc{\lowercase{#1}}\xspace}}
\tsc{WGM}
\tsc{QE}

\begin{document}

\let\WriteBookmarks\relax
\def\floatpagepagefraction{1}
\def\textpagefraction{.001}

\shorttitle{Efficiently Expanding Receptive Fields: Local Split Attention and Parallel Aggregation for Enhanced Large-scale Point Cloud Semantic Segmentation}
\shortauthors{Haodong Wang et~al.}

\title [mode = title]{Efficiently Expanding Receptive Fields: Local Split Attention and Parallel Aggregation for Enhanced Large-scale Point Cloud Semantic Segmentation}

\author[1]{Haodong Wang}
\fnmark[1]
\ead{21021210997@stu.xidian.edu.cn}
\credit{Formal analysis, Methodology, Software, Writing - Writing original draft}
\address[1]{Department of Remote Sensing Science and Technology, School of Electronic Engineering, Xidian University, 710071 Xi'an, China}

\author[2]{Chongyu Wang}
\fnmark[1]
\ead{chongyu@stu.xjtu.edu.cn}
\credit{Formal analysis, Software, Writing - Writing original draft}
\address[2]{School of Software Engineering, Xi'an Jiaotong University, 710049 Xi'an, China}

\author[1]{Yinghui Quan}
\ead{yhquan@mail.xidian.edu.cn}
\credit{Funding acquisition, Supervision, Writing - review and editing}

\author[2]{Di Wang}
\cormark[1]
\ead{diwang@xjtu.edu.cn}
\credit{Conceptualization, Funding acquisition, Methodology, Software, Supervision, Writing - review and editing}

\cortext[cor1]{Corresponding author}
\fntext[fn1]{The first two authors contributed equally to this work.}

\begin{abstract}
Expanding the receptive field in a deep learning model for large-scale 3D point cloud segmentation is an effective technique for capturing rich contextual information, which consequently enhances the network's ability to learn meaningful features. However, this often leads to increased computational complexity and risk of overfitting, challenging the efficiency and effectiveness of the learning paradigm. To address these limitations, we propose the Local Split Attention Pooling (LSAP) mechanism to effectively expand the receptive field through a series of local split operations, thus facilitating the acquisition of broader contextual knowledge. Concurrently, it optimizes the computational workload associated with attention-pooling layers to ensure a more streamlined processing workflow. Based on LSAP, a Parallel Aggregation Enhancement (PAE) module is introduced to enable parallel processing of data using both 2D and 3D neighboring information to further enhance contextual representations within the network. In light of the aforementioned designs, we put forth a novel framework, designated as LSNet, for large-scale point cloud semantic segmentation. Extensive evaluations demonstrated the efficacy of seamlessly integrating the proposed PAE module into existing frameworks, yielding significant improvements in mean intersection over union (mIoU) metrics, with a notable increase of up to 11\%. Furthermore, LSNet demonstrated superior performance compared to state-of-the-art semantic segmentation networks on three benchmark datasets, including S3DIS, Toronto3D, and SensatUrban. It is noteworthy that our method achieved a substantial speedup of approximately 38.8\% compared to those employing similar-sized receptive fields, which serves to highlight both its computational efficiency and practical utility in real-world large-scale scenes.

\end{abstract}
\begin{keywords}
	\sep Point Cloud
	\sep Semantic Segmentation
	\sep Local Split Attention
	\sep Parallel Aggregation
\end{keywords}

\maketitle

\section{Introduction}
The advancement of point cloud acquisition technology has precipitated a revolutionary transformation in the field of large-scale three-dimensional (3D) mapping of urban environments, with profound implications for a multitude of applications, including autonomous driving \citep{li2020deep}, urban planning \citep{urech2020point}, and environmental monitoring \citep{guo2020lidar}. The rich and high-resolution spatial information embedded in point cloud data offers invaluable insights for comprehensive scene understanding. Semantic segmentation, a process entailing the assignment of a semantic label to each point, represents a critical step in this process and has long been a focal point in the domains of remote sensing and computer vision \citep{guo2020deep}.

Early point cloud semantic segmentation methods frequently employed conventional machine learning techniques, utilizing manually crafted features and traditional classifiers such as random forests \citep{shotton2008semantic}. These methods demonstrated certain limitations when confronted with complex scenes. In recent years, with the rapid advancement of deep learning technology, deep learning-based methods have become the dominant approach in point cloud semantic segmentation. However, raw point clouds are typically irregularly sampled, unstructured, and unordered \citep{cui2021deep}, models for image processing like Convolutional Neural Networks (CNN) cannot be directly applied to point cloud data. Subsequently, projection-based methods \citep{Lawin_Danelljan_Tosteberg_Bhat_Khan_Felsberg_2017, Boulch_Saux_Audebert_2017, Tatarchenko_Park_Koltun_Zhou_2018, Lyu_Huang_Zhang_2022, zhang2023pointmcd}, which project point clouds into diverse two-dimensional (2D) representations, and voxel-based methods \citep{huang2016point,cciccek20163d,Riegler_Ulusoy_Geiger_2017,Wang_Huang_Shan_He_2018,LI2023103391}, which partition point clouds into fixed-size 3D grids, are developed. However, these methods entail considerable computational costs and may result in the loss of geometric structures. To address these limitations, the pioneering work of PointNet \citep{pointnet} was developed to directly process 3D points. Since then, numerous points-based complex neural networks have been proposed \citep{pointnet++,dgcnn,ldgcnn,pointnext}. These methods demonstrated satisfactory performance on small-scale point clouds. In recent years, there has been a notable shift in focus towards large-scale point cloud processing, driven by the rapid advancement of point cloud acquisition techniques and the digitization of numerous domains, including smart cities and natural resource management. Nonetheless, most classical deep learning networks are not readily applicable to large-scale point cloud semantic segmentation due to the intricate nature of their designs and the considerable computational costs involved. Consequently, a number of networks were designed with the specific objective of reducing computational cost and learning more representative features, particularly for large-scale point clouds \citep{randla, neiea}. This significantly accelerated the development of methods for the analysis of large-scale point clouds.

A fundamental aspect of deep learning-based point cloud semantic segmentation is the ability to sense local contextual information. In most existing networks, the point-wise features are learned by integrating neighboring features to enhance the model's capacity for perception \citep{Qiu_Anwar_Barnes_2021,randla,pointnext,neiea,lacv}. For instance, the K Nearest Neighbor (KNN) approach is frequently utilized to allocate surrounding information of a point. The spatial extent of the KNN, in turn, determines the receptive field of the model. In large-scale analysis, this stage is particularly crucial as objects are large in size and often span across a vast area. Thus, a large receptive field (e.g., a larger K value in KNN) is advantageous to encompass more neighboring information, which boosts the model's contextual comprehension. However, increasing the model's receptive field generally significantly increases the network's computational cost \citep{randla}. Moreover, an excessively large receptive field may impede the model's learning process to cause overfitting. Consequently, it is crucial to identify strategies for efficiently and effectively expanding the model's receptive field when dealing with large-scale scenes.

In order to address these issues, we introduce the local split attention mechanism in local feature learning. This strategy effectively enlarges the receptive field while minimizing the computational burden. Furthermore, we combine both 2D and 3D based KNN algorithms to further enlarge the receptive field by adding a parallel feature aggregation module. This module can be directly inserted into existing networks to enhance performance. Based on these designs, we introduce a new deep learning deep neural network called LSNet. Extensive experiments demonstrated its superior performance on three large-scale datasets. In summary, our main contributions are as follows:

\begin{enumerate}

	\item[$\bullet$] A Local Split Attention Pooling (LSAP) mechanism is designed to obtain a more extensive range of contextual information. In addition, this mechanism exhibits reduced computational costs. The effectiveness and efficiency of this mechanism have been demonstrated through rigorous experimentation.

	\item[$\bullet$] Base on LSAP, we further introduce a transferable Parallel Aggregation Enhancement (PAE) module. This module enriches local feature learning by incorporating both 2D and 3D neighboring information. The transferability and effectiveness of the module to existing networks was subjected to comprehensive testing.

	\item[$\bullet$] The integrated new network, LSNet, has demonstrated favorable performance on three large-scale benchmark datasets, including S3DIS, Toronto3D, and SensatUrban. Experiment results showed that it outperformed existing state-of-the-art networks.

\end{enumerate}\label{key}

\section{Related Work}
This section provides an overview of the principal techniques and associated developments in point cloud semantic segmentation. These encompass projection-based, voxel-based, and point-based methods. We also summarize the latest development of large-scale point cloud segmentation.

\subsection{Projection-based Methods}
Projection-based methodologies employ the projection of point clouds onto multiple 2D planes to generate synthetic images of various viewpoints. Subsequent processing then enables the acquisition of semantic segmentation labels for the point clouds using image processing networks. For example, \cite{Lawin_Danelljan_Tosteberg_Bhat_Khan_Felsberg_2017} projected the point cloud onto a set of synthetic images,  which are then used as input to the deep network. The resulting pixel-wise segmentation scores are re-projected into the point cloud. Similarly, \cite{Boulch_Saux_Audebert_2017} segmented point clouds by acquiring RGB and depth images from various perspectives. However, these multi-view methods are highly sensitive to occlusion. \cite{Tatarchenko_Park_Koltun_Zhou_2018} designed tangent convolutions to project local surface geometry onto tangent planes surrounding each point. EllipsoidNet \citep{Lyu_Huang_Zhang_2022} reduced overlap inside the points among points by projecting the point cloud onto ellipsoidal surfaces. However, projection-based methods inherently lose the geometric structure of point clouds. To solve this problem, Flattening-Net \citep{Zhang_Hou_Qian_Zeng_Zhang_He_2022} implicitly approximated a local smooth 3D-to-2D surface unfolding process, while preserving the geometric structure of point clouds. PointMCD \citep{zhang2023pointmcd} and PointVST \citep{Zhang_Hou_2022} augmented point-wise embeddings by capitalizing on 2D images derived from point cloud projection, subsequently employing cross-modal distillation and self-supervised learning. However, these methods require the implementation of preprocessing steps that are specifically designed and are limited in their ability to segment large-scale point clouds in complex environments.

\begin{figure*}[t]
	\begin{center}
		\includegraphics[width=1\textwidth]{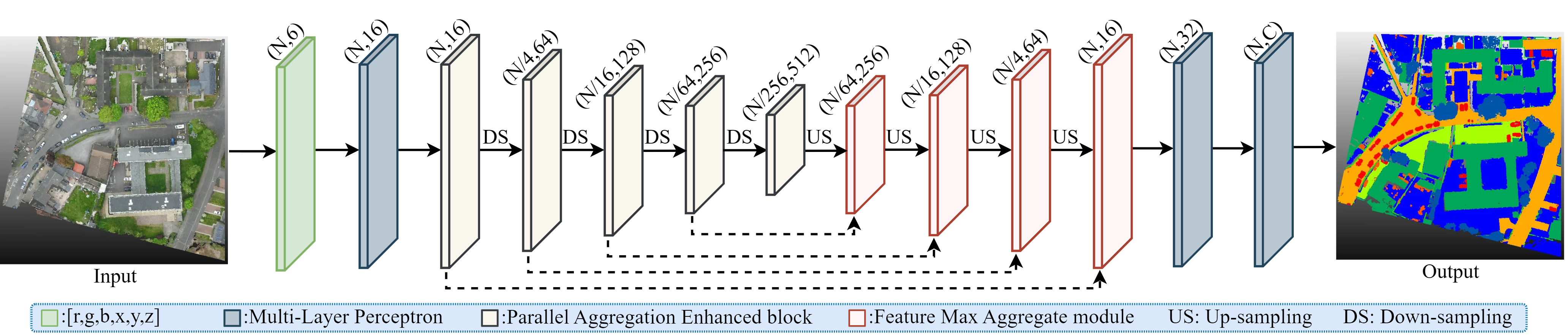}
	\end{center}
	\caption{The overall architecture of the proposed LSNet.}
	\label{fig:zero}
\end{figure*}

\subsection{Voxel-based Methods}
Voxel-based methods segment point clouds by dividing them into regular voxel grids for subsequent processing. Initially, point clouds were partitioned into uniform voxels where 3D convolutions were directly applied for semantic segmentation \citep{huang2016point, cciccek20163d}. However, increased voxel resolution escalated memory and computational demands to impractical levels. To mitigate this, OctNet \citep{Riegler_Ulusoy_Geiger_2017} utilized an octree to partition the point cloud into non-uniform voxels, allowing memory allocation and computation to concentrate on dense areas of the point cloud. MSNet \citep{Wang_Huang_Shan_He_2018} captured point cloud features across multiple scales and employs Conditional Random Fields (CRF) to ensure spatial consistency. \cite{Graham_Engelcke_Maaten_2018} introduced the submanifold sparse convolutional network (SSCN), which demonstrated remarkable efficacy in handling high-dimensional, sparse data, significantly enhancing precision and processing efficiency in 3D semantic segmentation. Cylinder3D \citep{Zhou_Zhu_Song_Ma_Wang_Li_Lin_2020} utilized a cylindrical partition based on cylindrical coordinates to address the issue of uneven point cloud density. (AF)2-S3Net \citep{Cheng_Razani_Taghavi_Li_Liu_2021} incorporated a multibranch attentive feature fusion module, adapting to different levels of sparsity within the point cloud. DRINet++ \citep{Ye_Wan_Xu_Cao_Chen_2021} conceptualized voxels as points to exploit the point cloud's sparsity and geometric features optimally. In more recent works, MVPNet \citep{LI2023103391} utilized a multi-scale voxel fusion module coupled with a geometric self-attention module for segmentation of urban scene point clouds. Nevertheless, reduced voxel resolutions may compromise detailed spatial features, and scaling up voxel resolution in large-scale point clouds necessitates significant increases in memory and computational resources, constraining the use of voxel-based methods for large-scale point cloud semantic segmentation.

\subsection{Point-based Methods}
The point-based approach directly takes point cloud data as input. Early on, PointNet \citep{pointnet} emerged as a milestone in point cloud deep learning methods, capable of directly handling point cloud data and capturing global information of points. However, it lacks the capability to learn local features, limiting its segmentation performance. To address this limitation, \cite{pointnet++} proposed PointNet++ based on PointNet, which effectively solves the two problems of dividing local point clouds and extracting local features of point clouds through three parts: a sampling layer, a grouping layer, and a feature extraction layer. Inspired by these developments, many efforts have been made to enhance the performance of point cloud semantic segmentation. For instance, DGCNN \citep{dgcnn} addressed the issue of poor local feature acquisition by introducing edge convolution modules, which are stacked or recurrently applied to obtain global features. However, it overlooks the directional information between points, resulting in some information loss. To address this issue, \cite{ldgcnn} proposed the Link Dynamic Graph Convolutional Neural Network (LDGCNN). It not only eliminates the spatial transformation network with too many parameters but also aggregates the hierarchical features of dynamic graphs from different layers by adding skip links, enabling the model to learn effective edge vector features from the features. KPConv \citep{kpconv} proposed the use of kernel points that weigh features of different dimensions based on distance. However, it is necessary to adapt the size of the kernel points to accommodate the inherent variability in point cloud data. In general, point-based methods are regarded as more effective than projection- and voxel-based techniques for the comprehensive exploitation of the 3D data embedded in point clouds.

\subsection{Large-scale Point Cloud Segmentation}
In recent years, \cite{randla} proposed a large-scale point cloud segmentation method called RandLA-Net. By employing random sampling and local feature aggregation, it significantly improves the processing speed, accuracy, and robustness of point cloud segmentation. Moreover, the emergence of RandLA-Net has prompted increased attention to the effective acquisition of local context information. For example, \cite{Qiu_Anwar_Barnes_2021} enhanced the local context of points by introducing a bilateral block, which leverages both geometric and semantic features to extract valuable neighboring point information. PointNeXt \citep{pointnext} improved and optimized PointNet by redesigning efficient training strategies and model scaling. It introduced techniques such as multi-scale feature extraction, adaptive aggregation modules, and local relationship modeling, enhancing the network's performance and generalization capabilities. NeiEA-Net \citep{neiea} focused on optimizing local features by adaptively aggregating features from different scales using neighborhood feature aggregation modules. This approach allows for the reduction of redundant information and the effective learning of local details. LACV-Net \citep{lacv} addressed the issue of local perceptual blurriness caused by the KNN algorithm through three designs: local adaptive feature enhancement modules, aggregation loss functions, and local aggregation descriptor vector synthesis modules. This enables it effectively captures global features scattered across large-scale scenes. These works have all demonstrated that the key to improving the segmentation performance of point clouds in complex scenes lies in the ability to effectively capture local context information.

Overall, in the domain of large-scale point cloud segmentation, the integration of local contextual information derived from neighboring points typically requires a broad receptive field to capture meaningful geometric details within the point clouds. However, this expansion of the receptive field may result in a reduction in computational efficiency and an increased risk of overfitting.

\section{Methodology}
The proposed LSNet consists of three main components: the \textbf{Local Split Attention Pooling} (LSAP) mechanism, the \textbf{Parallel Aggregation Enhancement} (PAE) module, and the \textbf{Feature Max Aggregated} (FMA) module, which are described in details in this section.

\begin{figure*}[t]
	\begin{center}
		\includegraphics[width=1\textwidth]{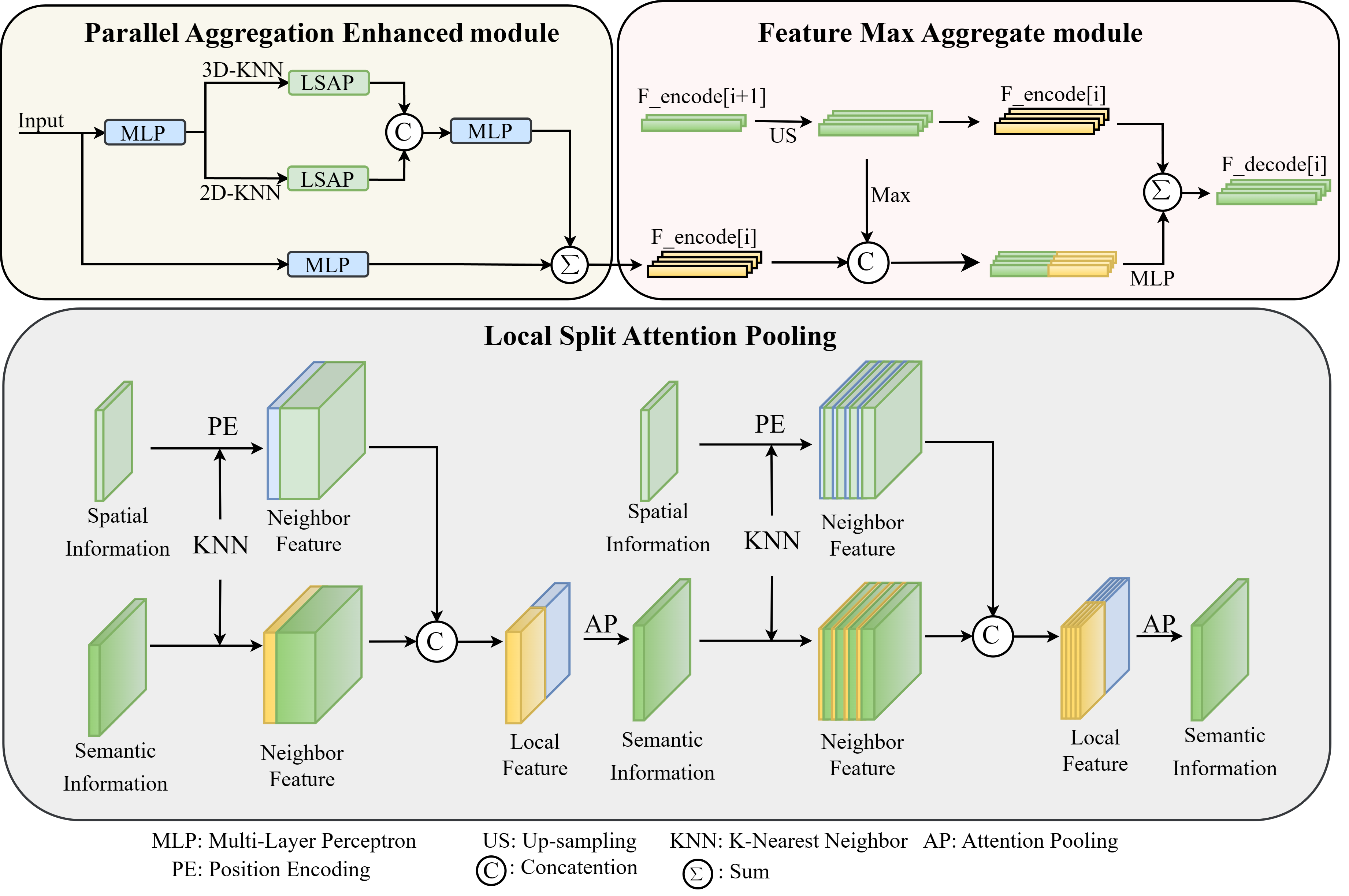}
	\end{center}
	\caption{Key components of the proposed LSNet. The yellow panel displays the parallel aggregation enhancement module, which is used as the main framework for the encoding layer. The pink panel diagram shows the feature max aggregate module, which is responsible for feature aggregation between point features at various scales. The gray panel shows the details of the local split attention pooling mechanism.}
	\label{fig:one}
\end{figure*}
\subsection{Network Framework}\label{nf}

Figure \ref{fig:zero} illustrates the overall architecture of the proposed LSNet, while detailed components are elaborated in Figure \ref{fig:one}. At the encoding layer, the model uses the PAE module with the LSAP mechanism to learn the local features of points at different resolutions through layer-by-layer stacking. At the decoding layer, the model uses the FMA module to aggregate the upsampled features with the features in the same encoding layer. Finally, the model obtains the semantic segmentation results of the original scale layer through fully connected layers.

\subsection{Local Split Attention Pooling Mechanism}\label{lap}
We propose a mechanism called Local Split Attention Pooling (LSAP). This mechanism essentially splits the current features twice before passing them to the attention pooling layer in a skip fashion, rather than passing them all at once, to reduce computational cost and also improve the network's ability to extract useful information from local features. The details of this idea are shown on the gray panel in Figure \ref{fig:one}. The workflow of the LSAP can be seen in the Algorithm \ref{alg:alg1}.

\renewcommand{\algorithmicrequire}{\textbf{Input:}}
\renewcommand{\algorithmicensure}{\textbf{Output:}}

\begin{algorithm}[h]
	\linespread{1.15}\selectfont.
	\caption{Local Split Attention Pooling (LSAP)}
	\label{alg:alg1}
	\begin{algorithmic}[1]
		\Require
		Semantic features \( F = \{f_i\}_{i=1}^{N} \) and point spatial coordinates \( P = \{p_i\}_{i=1}^{N} \).
		\Ensure
		The features that include the information of neighboring points \( F_{\mathrm{att2}} = \{f_i\}_{i=1}^{N} \)
		\State $\text{neigh\_idx}(p_i) = \mathrm{KNN}(P, p_i, k)$
		\State $\text{split\_idx1}(p_i)=\text{neigh\_idx}(p_i)[0:s_1]$\Comment {first split}
		\State $P_{\mathrm{RPPF}}=\{ {r_i^k} \}=PE(p_i,\text{split\_idx1}(p_i))$ \Comment {positional encoding}
		\State $F_{\mathrm{ne}}=\{ {f_i^k} \}=Gather(f_i,\text{split\_idx1}(p_i))$\Comment {gather features of neighbor points}
		\State $F_{\mathrm{con}}=concat(F_{\mathrm{ne}},P_{\mathrm{RPPF}})$
		\State $F_{\mathrm{att1}} =\{ {f_i^{att1}} \}_{i=1}^{N}= att(F_{\mathrm{con}})$  \Comment {attention pooling}
		\State $\text{split\_idx2}(p_i)=\text{neigh\_idx}(p_i)[0::s_2]$\Comment {second split}
		\State $P_{\mathrm{RPPF}}=\{ {r_i^k} \}=PE(p_i,\text{split\_idx2}(p_i))$
		\State $F_{\mathrm{ne}}=\{ {f_i^k} \}=Gather(f_i^{att1},\text{split\_idx2}(p_i))$
		\State $F_{\mathrm{con}}=concat(F,P_{\mathrm{RPPF}})$
		\State $F_{\mathrm{att2}} = att(F_{\mathrm{con}})$
	\end{algorithmic}
\end{algorithm}

Firstly, we take the semantic features \( F = \{f_i\}_{i=1}^{N} \) and point spatial coordinates \( P = \{p_i\}_{i=1}^{N} \) as inputs to the LSAP block. Then, the KNN algorithm is used to search for the neighboring points index of each point (i.e., $\text{neigh\_idx}(p_i)$). It is worth noting that the indexes of neighboring points in {$\text{neigh\_idx}(p_i)$} are arranged in ascending order of their distance from the central point, and the KNN algorithm in the module searches for neighboring points according to the standards specified by each branch in the PAE module. Then the LASP block performs a splitting operation on the neighboring points index:
\begin{equation}\label{eq0}
	\text{split\_idx1}\left( p_i\right)=\text{neigh\_idx}\left( p_i\right)\left[0:s_1\right]
\end{equation}
\noindent
where $s_1$ is a hyperparameter indicating that we only need to do subsequent computations for the top $s_1$ neighbor points. We always let $s_1 = \frac{1}{4}k$. $\text{split\_idx1}$ is the neighbor point index that participates in subsequent operations. The primary objective of the first split is to reduce the computational cost of the subsequent attention pooling step. Since the $\text{neigh\_idx}$ are arranged in ascending order of their distance from the central point, the first split selects the $s_1$ nearest points to the central point. This approach also ensures that fine-grained features are captured. Then we reference the Relative Point Position Encoding (RPPE) method from RandLA-Net \citep{randla}, which explicitly embeds the spatial coordinates of neighboring points. This ensures that the central point is always aware of the relative spatial positions of its neighboring points, facilitating the network's learning of complex structures. The RPPE expression for $P_{\mathrm{RPPF}}=\{ {r_i^k} \}$ is:

\begin{equation}
	\begin{split}
		\{r_i^k\} = MLP\left( p_i \oplus p_i^k \oplus \left( p_i - p_i^k \right) \oplus \left\| p_i - p_i^k \right\| \right)
	\end{split}
\end{equation}

\begin{equation}
	\{p_i^k\} = \{p_j \mid j \in \text{split\_idx1}(p_i)\}
\end{equation}

\noindent
where $\{p_i^k\} $ is the set of coordinates of the neighboring points of $p_i$ after splitting.

Then, the features of the neighbors of each point $F_{\mathrm{ne}}$ are collected. Subsequently, the features $F_{\mathrm{ne}}$ and $P_{\mathrm{pe}}$ are concatenated as $F_{\mathrm{con}}=concat(F_{\mathrm{ne}},P_{\mathrm{pe}})$. Attention pooling is then performed on the local feature  $F_{\mathrm{con}}=\{ {f_i^k} \}$, i.e., $F_{\mathrm{att2}} = att(F_{\mathrm{con}})$ expressed as:

\begin{equation}\label{eq2}
	K = {\mathop{\rm Softmax}\nolimits} \left( {\rm MLP}  \left(  F_{\mathrm{con}}\right),axis=-2\right)
\end{equation}

\begin{equation}\label{eq2}
	V =   {\rm MLP}  \left(  F_{\mathrm{con}}\right)
\end{equation}

\begin{equation}\label{eq4}
	F_{\mathrm{att1}}= {\rm MLP}\left( {\mathop{\rm Sum}\nolimits}\left( {K \cdot V} \right), axis=-2 \right)+F
\end{equation}

\noindent
In this operation, two multi-layer perceptrons (MLPs) are employed to learn $F_{\mathrm{con}}$ in a separate manner. Subsequently, a softmax calculation is performed on one of the branches, designated as $K$, to derive the attention scores for each neighboring feature. Thereafter, $K$ and $V$ are calculated by multiplying the points, followed by accumulation. It is noteworthy that residual connections are employed to enhance the stability of the network. Ultimately, the feature $F_{\mathrm{att1}}$ is obtained, which contains information from neighboring points.

As described in Algorithm \ref{alg:alg1}, the above operation performs the first round of local split attention pooling for LSAP. In order to expand the receptive field of the network, we perform a second round of local split attention pooling. The difference is that in the second split, we take a point every $s_2$ neighboring points to participate in the subsequent operation, which is expressed as:

\begin{equation}\label{eq5}
	\text{split\_idx2}\left( p_i\right)=\text{neigh\_idx}\left( p_i\right)\left[0::s_2\right]
\end{equation}
\noindent
We always let $s_2 = 4$. Accordingly, the subsequent operations are identical to those of the initial round of splitting. Subsequently, following the second attention pooling, the final features are obtained. It is assumed that neighboring points often share similar semantic features. Therefore, in the second split, every $s_2$-th neighboring point is selected for subsequent computations, thereby achieving a larger receptive field with lower computational cost. By employing these two splits, the LSAP mechanism strikes a balance between a large receptive field and computational cost, while also preserving the fine-grained features of the point cloud.

\begin{figure*}[h]
	\begin{center}
		\includegraphics[width=0.7\textwidth]{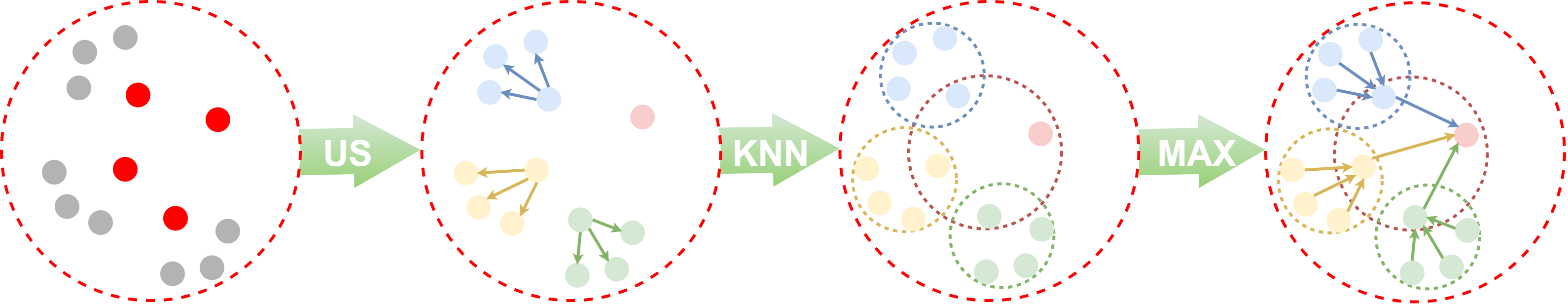}
	\end{center}
	\caption{Upsampling and max pooling components. Red point: low-resolution layers features, gray point: high-resolution layers features, US: upsampling, Max: max pooling, KNN: K-Nearest Neighbor}
	\label{fig:three}
\end{figure*}

\subsection{Parallel Aggregation Enhancement Module}\label{pae}
Each encoder layer comprises a PAE module. The purpose of this module is to optimize the aggregation of local features for each point at different resolutions. In contrast to previous works that employed solely the 3D-KNN algorithm to extract local features for each point, our approach integrates the 2D-KNN with 3D-KNN to concurrently capture local features. The rationale is based on the observation that in large-scale point clouds, such as those representing urban scenes, the objects (i.e., buildings and trees) themselves primarily have an elongated shape along the vertical dimension, while exhibiting a planar distribution across a vast region. It would be optimal to augment the contextual information associated with the vertical structure. Accordingly, both 2D and 3D KNN are introduced as means of concurrently obtaining local features. Specifically, the 2D-KNN is achieved by setting the Z value of the point coordinates to zero. The parallel structure enables the model to simultaneously obtain two sets of distinct local features, thereby facilitating a more comprehensive learning process and enhancing its capacity to learn complex objects.

As illustrated in the upper left quadrant of Figure \ref{fig:one}, the initial step involves feeding the supplied input features into the MLP to augment the number of channels for the original features. The model is divided into two parallel branches, with each branch employing a different search method: 2D-KNN and 3D-KNN. Subsequently, the local features obtained from each branch are conveyed to the LSAP block for a local split attention pooling operation. This operation is responsible for the collection of effective point feature vectors derived from the local features. To reduce the learning cost of the model, the number of channels for each LSAP block is reduced by half of the original. Subsequently, the two sets of feature vectors are concatenated and sent to MLP for learning, thereby obtaining point feature vectors containing both types of features. Ultimately, the point feature vector is added to the original feature vector to form a residual block, thus facilitating the learning of the model and enhancing its stability and reliability. %

\subsection{Feature Max Aggregated Module}\label{mar}
This module functions at the decoding layer, where it aggregates point features at varying resolutions. The objective is to ensure the preservation of the most useful features learned from each layer to the greatest extent. As illustrated in the upper right quadrant of Figure \ref{fig:one}, the input to this module is the point feature set, denoted by ${F_{encode}}\left[ i \right]$, for each scale reserved by the coding layer. Firstly, the resolution of ${F_{encode}}\left[ i+1 \right]$ is adjusted to align with that of ${F_{encode}}\left[ i \right]$ through upsampling. It should be noted, however, that the point cloud sampling method via KNN cannot be sampled at regular intervals in the same way as an image can be upsampled. As illustrated in Figure \ref{fig:three}, due to the unordered nature of point cloud data, stacking occurs after upsampling, and some points are unable to effectively propagate to the high-resolution layer. Consequently, after aggregating the upsampled features with those of the neighboring points, a max pooling operation was performed to focus on the most prominent features. The specific operation is illustrated by the following expression:

\begin{equation}\label{eq7}
	F_{m}[i] = \text{Gather}\left(\text{US}\left(F_{decode}[i + 1]\right), \text{neigh\_idx}\right)
\end{equation}

\begin{equation}\label{eq6}
	\begin{split}
		F_{decode}[i] = & \text{MLP} \left( \text{MAX}\left(F_{m}[i]\right); F_{encode}[i] \right) \\
		& + F_{encode}[i]
	\end{split}
\end{equation}

\noindent
where ${F_{encode}}\left[ i \right]$ is the feature matrix saved by each encoding layer, and ${F_{decode}}\left[ i \right]$ is the feature matrix saved by each decoding layer. $F_{decode}[i + 1]$ represents the previous decoding layer. Firstly, the previous decoding layer is upsampled, then aggregated with the features of neighboring points (gather), and subsequently max-pooled to yield $F_m$. Thereafter, $F_m$ is concatenated with the corresponding encoding layer features, $F_{encode}[i]$, which are then fed into an MLP for feature learning. Additionally, residual blocks are incorporated to guarantee model stability and to prevent any adverse effects from invalid low-resolution feature matrix data.

\section{Experiments and Results}
In this section, we conduct experiments on three large-scale datasets to validate the effectiveness of LSNet and perform ablation studies on the proposed LSAP mechanism, PAE, and FMA modules.

\subsection{Experimental Details}

We train 100 cycles on a single GeForce RTX 3090 GPU, and the batch size ranges from 4 to 16, depending on the number of input points for different datasets (ranging from 16 $\times$ 1024 to 64 $\times$ 1024). In addition, the Adam optimizer \citep{adam} is utilized to minimize the total loss. The learning rate is initialized at 0.01 and decays at a rate of 0.95 after each cycle. And we set the hyperparameter $s_1 = \frac{1}{4}k$ and $s_2=4$, where $k$ is the number of neighboring points in the KNN algorithm. The weighted cross-entropy, as utilized in RandLA-Net \citep{randla}, is also employed as the loss function. We used ubuntu18.04 to implement the following methods on Python and Tensorflow \citep{tf} platforms.

To quantitatively analyze the performance of the LSNet, we utilize common metrics including intersection over union (IoU), mean IoU (mIoU), and overall accuracy (OA) as evaluation metrics.

\subsection{Experimental Datasets}
To verify the effectiveness of our approach, we chose three widely-used and representative datasets that cover various real-world scenarios.

\begin{figure}
	\begin{center}
		\includegraphics[width = 0.5\columnwidth]{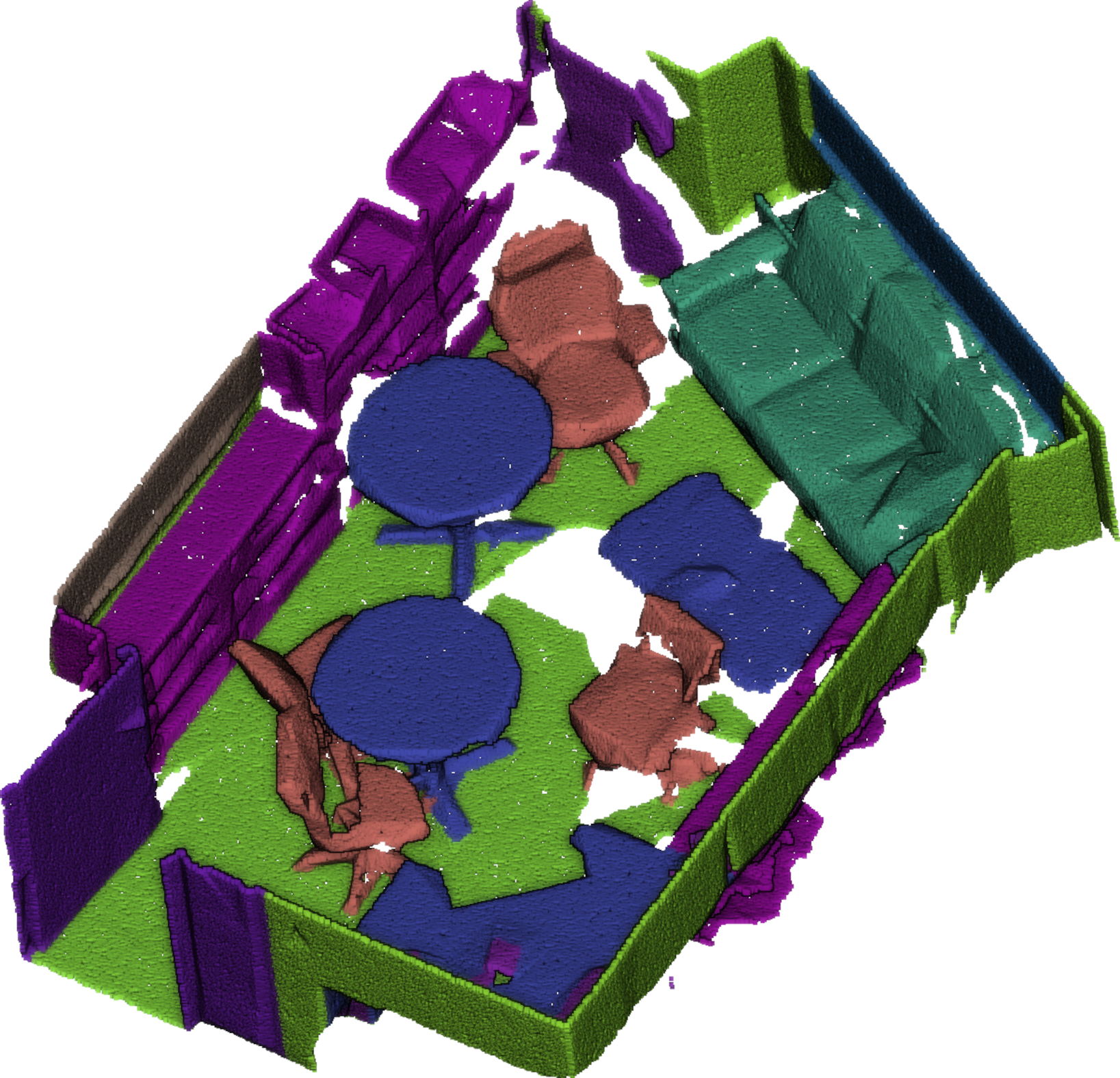}
	\end{center}
	\caption{Example view of D3DIS dataset. Each semantic category is colored randomly.}
	\label{s3disori}
\end{figure}

\begin{figure}
	\begin{center}
		\includegraphics[width = 1\columnwidth]{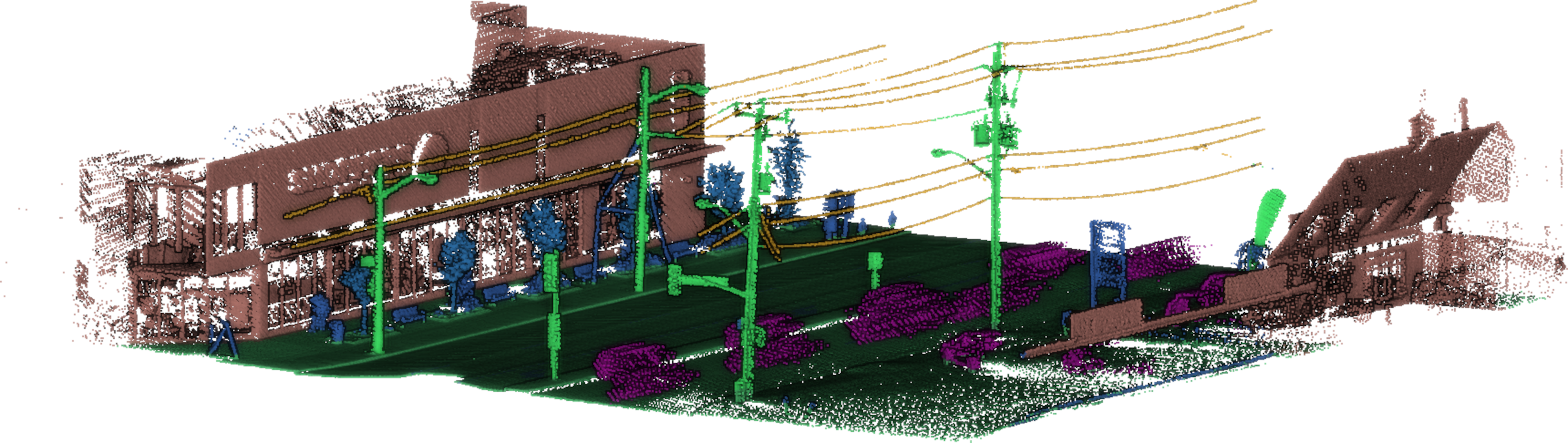}
	\end{center}
	\caption{Example view of Toronto3D dataset. Each semantic category is colored randomly.}
	\label{t3dori}
\end{figure}

\subsubsection{S3DIS Dataset}
The Stanford University Large 3D Indoor Space (S3DIS) dataset \citep{s3dis} comprises indoor 3D point cloud data collected with a Matterport camera (Figure \ref{s3disori}). The dataset comprises six subregions, 13 labels, and 11 scenarios. The scene encompasses a multitude of architectural elements, including offices, meeting rooms, corridors, auditoriums, open spaces, lobbies, lounges, storerooms, copy rooms, storage rooms, and bathrooms. The quantity of point cloud data present in each room is contingent upon its dimensions. The majority of rooms exhibit a point count ranging from 500,000 to 250,000. To evaluate our approach, we employ a sixfold strategy, as outlined in PointNet \citep{pointnet}.

\subsubsection{Toronto3D Dataset}
The Toronto3D dataset \citep{toronto} is a large-scale urban outdoor point cloud dataset obtained for semantic segmentation using the Teledyne Optech Maverick2 on-board Mobile Laser Scanning (MLS) system in Toronto, Canada. The dataset was collected on Avenue Road in Toronto, Canada, encompassing approximately 1 km of road (Figure \ref{t3dori}). The dataset comprises 78.3 million points and eight object classes, each with a unique tag. The dataset can be divided into four distinct parts. L001, L002, L003, and L004 each encompass an area of approximately 250 meters. L002 is employed for the purpose of verifying the accuracy of semantic segmentation tasks.

\begin{figure}
	\begin{center}
		\includegraphics[width = 1\columnwidth]{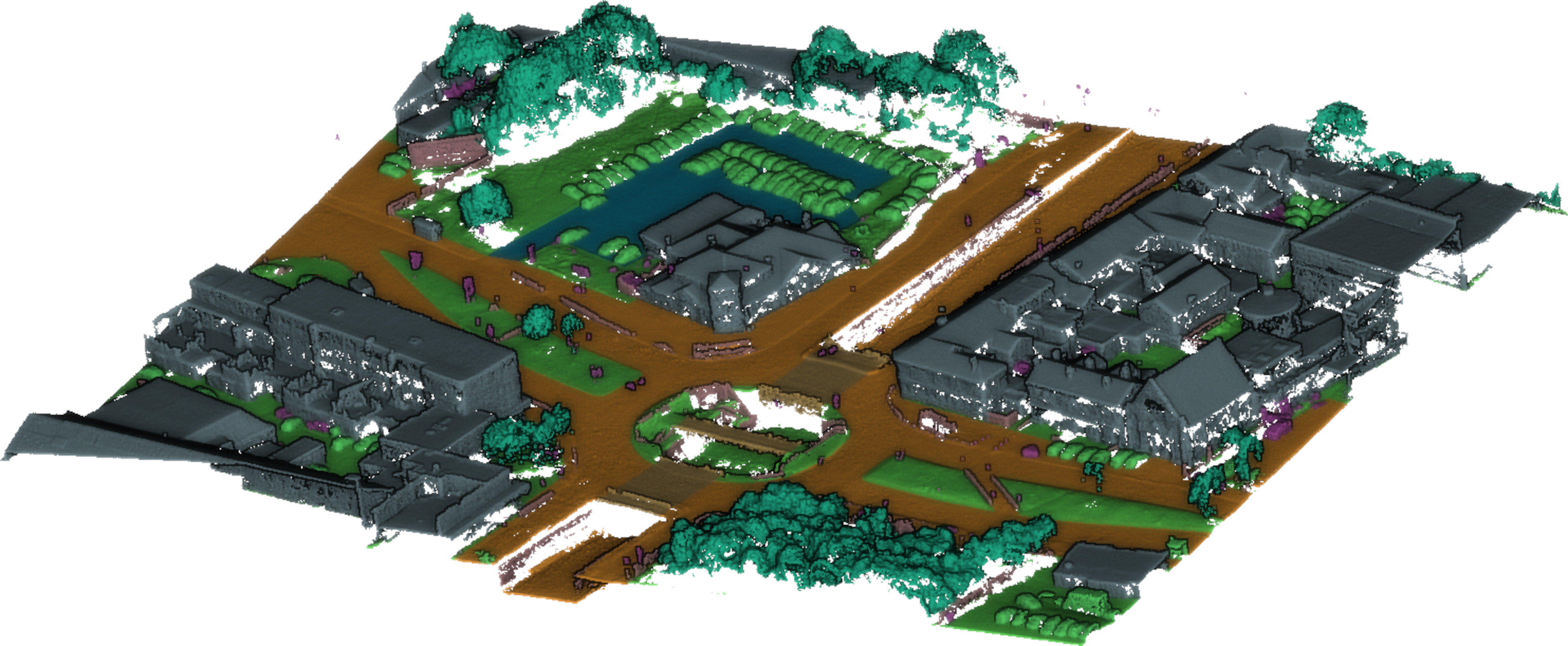}
	\end{center}
	\caption{Example view of SensatUrban dataset. Each semantic category is colored randomly.}
	\label{sensatori}
\end{figure}

\begin{table}
	\begin{center}
		\label{tab}
		\resizebox{\columnwidth}{!}{%
			\begin{tabular}{lcc}

				\hline
				Method                        & \rotatebox{0}{OA ($\% $)}        & \rotatebox{0}{mIoU ($\% $)}      \\
				\hline
				PointNet \citep{pointnet}     & 78.6                             & 47.6                             \\
				PointNet++ \citep{pointnet++} & 81.0                             & 54.5                             \\
				DGCNN \citep{dgcnn}           & 84.1                             & 56.1                             \\
				KPConv \citep{kpconv}         & -                                & 70.6                             \\
				ShellNet  \citep{shellnet}    & 87.1                             & 66.8                             \\
				PointWeb \citep{pointweb}     & 87.3                             & 66.7                             \\
				SPGraph  \citep{ssp+spg}      & 87.9                             & 68.4                             \\
				PointASNL \citep{pointasnl}   & 88.8                             & 68.7                             \\
				SCF-Net  \citep{scf}          & 88.4                             & 71.6                             \\
				CBL \citep{cbl}               & 88.9                             & 72.2                             \\
				PointNeXt-L \citep{pointnext} & 89.7                             & 73.9                             \\
				FA-ResNet \citep{fa}          & 88.8                             & 71.7                             \\
				LG-Net \citep{large}          & 88.3                             & 70.8                             \\
				LACV-Net \citep{lacv}         & 88.3                             & 70.8                             \\

				\hline
				RandLA-Net \citep{randla}     & 88.0                             & 70.0                             \\
				\textbf{+ PAE}                & \textcolor{blue}{88.5}           & \textcolor{blue}{72.5}           \\

				\hline
				NeiEA-Net \citep{neiea}       & 89.3                             & 72.9                             \\
				\textbf{+ PAE}                & \textcolor{blue}{89.6}           & \textcolor{blue}{73.5}           \\

				\hline
				LSNet                         & \textcolor{black}{\textbf{90.3}} & \textcolor{black}{\textbf{74.0}} \\
				\hline
			\end{tabular}
		}
	\end{center}
	\caption{S3DIS semantic segmentation results. Blue font color indicates improved results by adding our PAE module to baseline networks. The best result is marked in bold.}
	\label{tab:two}
\end{table}

\begin{figure*}
	\begin{center}
		\includegraphics[width = 0.8\textwidth]{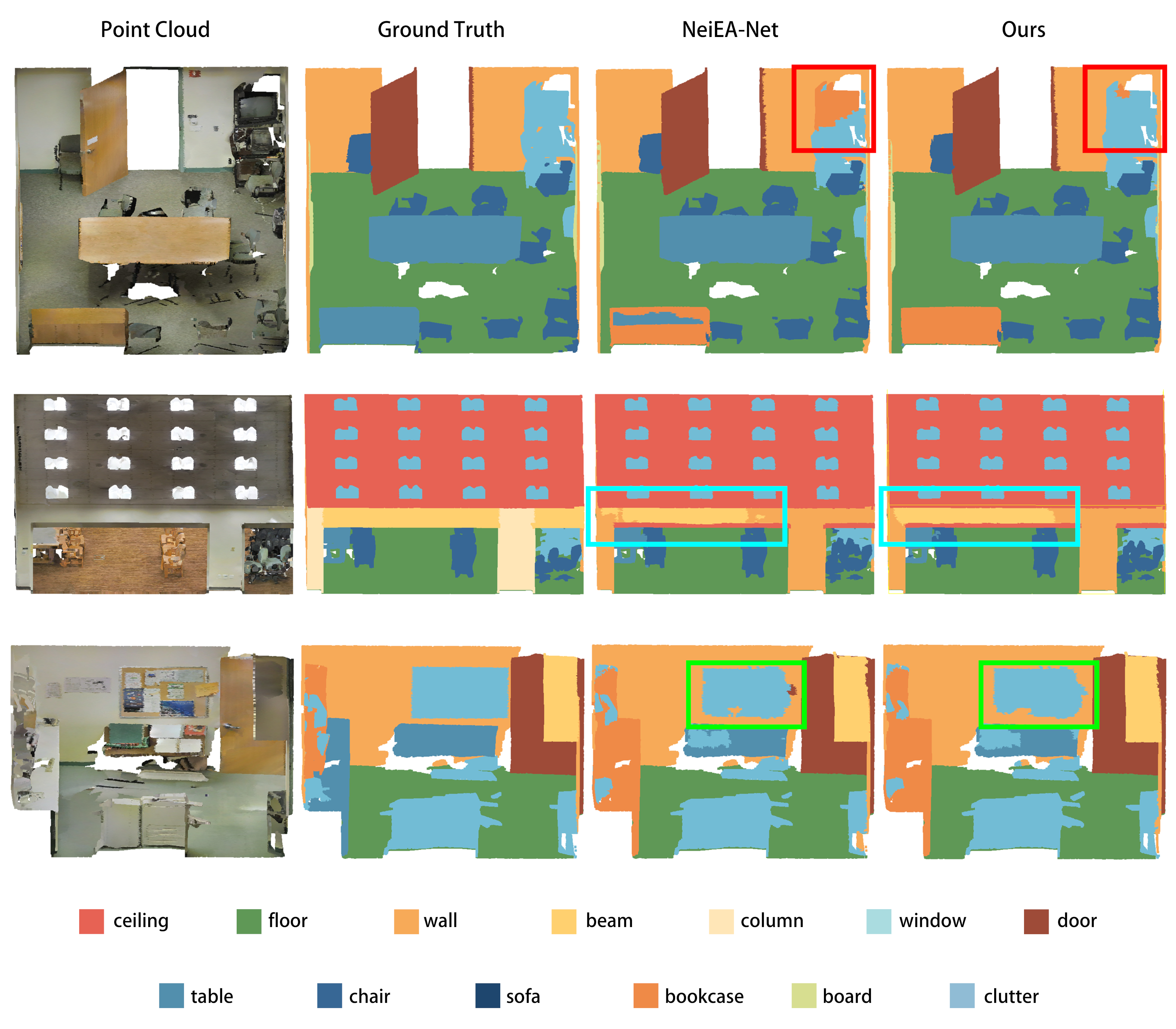}
	\end{center}
	\caption{Visualization of semantic segmentation result of S3DIS. The rectangles indicate regions where our method achieved superior results compared to others.}
	\label{fig:s3dis}
\end{figure*}

\subsubsection{SensatUrban Dataset}
The SensatUrban dataset \citep{sensaturban} comprises photogrammetric point cloud data collected over a 7.6-square-kilometre area of three British cities (Birmingham, Cambridge, and York) using a fixed-wing UAV. The dataset comprises approximately 3 billion points, each with detailed semantic annotation that classifies it into one of 13 different semantic categories (Figure \ref{sensatori}). The primary semantic categories are ground, building, and vegetation, which comprise over 70\% of the total points. In contrast, bicycle, rail, and other categories account for only about 0.025\% of the total points, illustrating an extremely unbalanced distribution of different semantic categories in the actual scene.

\subsection{Data Preprocessing}
In the data preprocessing stage, grid sampling was performed on the raw data.
For the S3DIS dataset, we adhere to the original dataset distribution before the experiment and implement the 6-fold cross-validation method. For each round of cross-validation, we designated five areas as the training set and the remaining one area as the test set. We conducted 0.04 m grid sampling during preprocessing. In the case of the Toronto3D dataset, the validation set was designated as L002, while the remaining datasets were assigned to the training set. During the preprocessing stage, a grid sampling method was employed at an interval of 0.06 m. The SensatUrban dataset comprises six test sets and four validation sets, with the remainder designated as training sets. A grid sampling of 0.2 m was employed for the large-scale urban datasets. In order to create the most efficient receptive field for the model, 16,384 neighborhood points were selected as input points based on the center points.

\begin{table*}[]
	\begin{center}
		\resizebox{\textwidth}{!}{
			\begin{tabular}{lccccccccccc}
				\hline
				Method                      & OA (\%)                          & mIoU (\%)                        & IoUs (\%)     &               &                                  &                                  &                                 &                                  &               &                                  \\ \cline{4-11}
				                            &                                  &                                  & road          & road m.       & natural                          & build.                           & util.l.                         & pole                             & car           & fence                            \\ \hline

				ResDLPS-Net \citep{resdlps} & 96.4                             & 80.2                             & 95.8          & 59.0          & 96.1                             & 90.9                             & 86.8                            & 79.9                             & 89.4          & 43.3                             \\
				BAAF-Net \citep{baaf}       & 94.2                             & 81.2                             & 96.8          & 67.3          & 96.8                             & 92.2                             & 86.8                            & 82.3                             & 93.1          & 34.0                             \\
				BAF-LAC \citep{baf}         & 95.2                             & 82.2                             & 96.6          & 64.7          & 96.4                             & 92.8                             & 86.1                            & 83.9                             & \textbf{93.7} & 43.5                             \\
				RG-GCN \citep{rg}           & 96.5                             & 74.5                             & \textbf{98.2} & 79.4          & 91.8                             & 86.1                             & 72.4                            & 69.9                             & 82.1          & 16.0                             \\

				LACV-Net \citep{lacv}       & 97.4                             & 82.7                             & 97.1          & \textbf{66.9} & 97.3                             & 93.0                             & 87.3                            & 83.4                             & 93.4          & 43.1                             \\

				\hline
				RandLA-Net \citep{randla}   & 94.3                             & 81.7                             & 96.6          & 64.2          & 96.9                             & 94.2                             & \textbf{88.0}                   & 77.8                             & 93.3          & 42.8                             \\
				\textbf{+ PAE}              & \textcolor{blue}{96.2}           & \textcolor{blue}{82.4}           & 96.5          & 64.1          & \textcolor{blue}{97.2}           & \textcolor{blue}{94.8}           & 87.3                            & \textcolor{blue}{82.6}           & 93.1          & \textcolor{blue}{43.5}           \\
				\hline
				NeiEA-Net \citep{neiea}     & 97.0                             & 82.6                             & 97.1          & \textbf{66.9} & 97.3                             & 93.0                             & 87.3                            & 83.4                             & 93.4          & 43.1                             \\
				\textbf{+ PAE}              & \textcolor{blue}{97.2}           & \textcolor{blue}{83.0}           & 96.6          & 66.6          & \textcolor{blue}{97.5}           & \textcolor{blue}{94.5}           & \textcolor{blue}{\textbf{88.0}} & \textcolor{blue}{83.5}           & 93.4          & \textcolor{blue}{44.5}           \\

				\hline
				LSNet                       & \textcolor{black}{\textbf{97.5}} & \textcolor{black}{\textbf{83.4}} & 96.8          & 65.5          & \textcolor{black}{\textbf{97.8}} & \textcolor{black}{\textbf{95.3}} & \textcolor{black}{87.8}         & \textcolor{black}{\textbf{84.7}} & 93.6          & \textcolor{black}{\textbf{45.1}}
				\\ \hline
			\end{tabular}
		}
	\end{center}
	\caption{Toronto3D semantic segmentation results. Blue font color indicates improved results by adding our PAE module to baseline networks. The best result is marked in bold.}
	\label{tab:three}
\end{table*}

\begin{figure*}
	\begin{center}
		\includegraphics[width=0.7\textwidth]{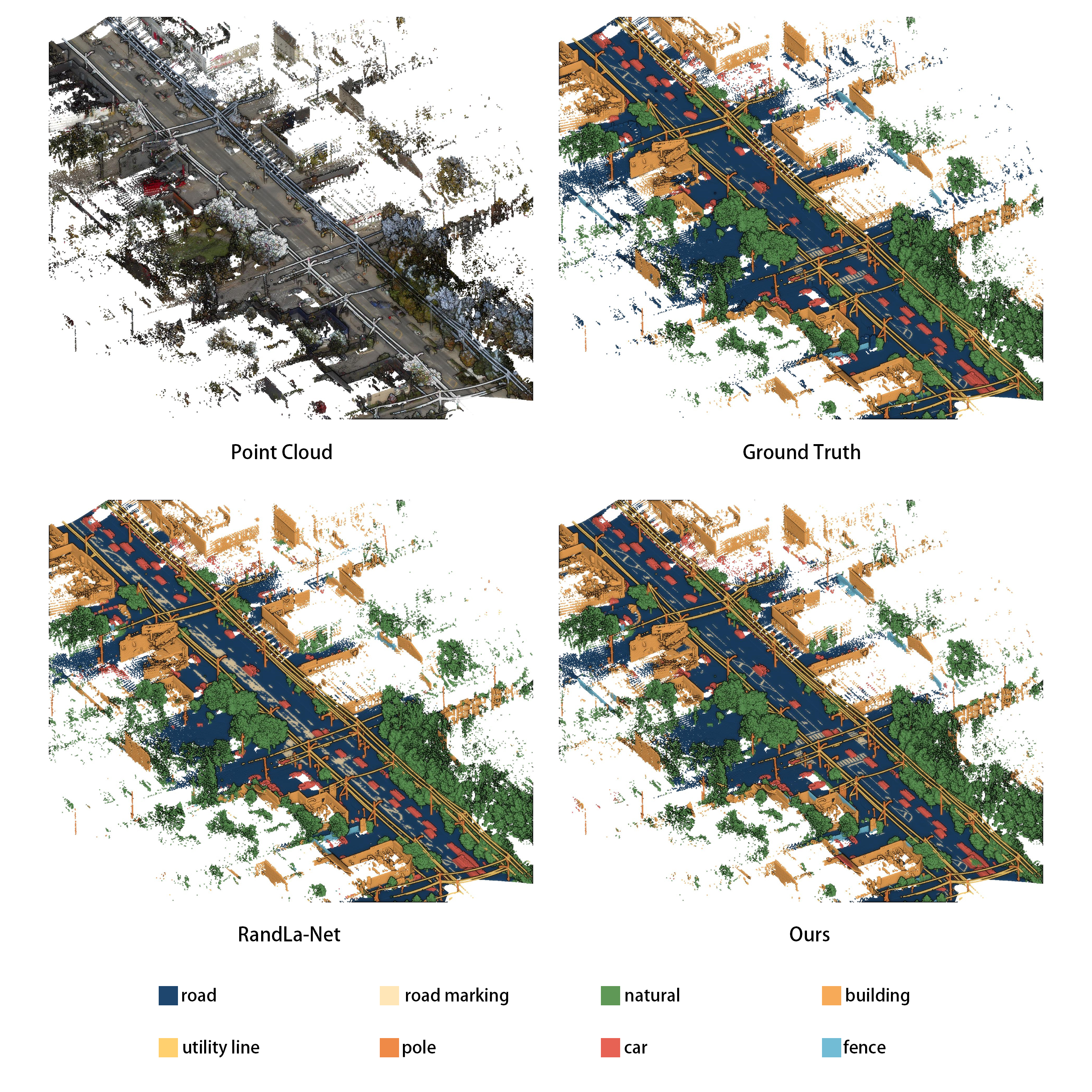}
	\end{center}
	\caption{Visualization of semantic segmentation result of Toronto3D.} %
	\label{fig:five}
\end{figure*}

\subsection{Semantic Segmentation Results}
The semantic segmentation performance of LSNet was evaluated on three large-scale datasets, namely S3DIS, Toronto3D, and SensatUrban. Results of the proposed LSNet were compared with other networks. Furthermore, to assess the efficacy of the devised PAE module with LSAP mechanism and parallel aggregate enhancement, a specific evaluation was partitioned into two categories: the baseline configuration and the augmented PAE module (baseline \textbf{+ PAE}). The first category comprises the original network, which employs solely 3D-KNN local feature learning without local split attention. The second category encompasses the network model with an integrated PAE module. The selected baseline networks are two prominent networks for large-scale point cloud segmentation: RandLA-Net \citep{randla} and NeiEA-Net \citep{neiea}.

\subsubsection{Evaluation on S3DIS}
Table \ref{tab:two} depicts the segmentation result of LSNet on S3DIS in comparison with other networks. In general, our network demonstrates clear superiority to existing methods in terms of OA ($90.3\%$) and mIoU ($74.0\%$). Furthermore, the incorporation of our transferable PAE module into existing networks resulted in notable improvements in performance, with an increase of up to 2.5\% in mIoU. This evidence supports the effectiveness of our approach. Figure ~\ref{fig:s3dis} illustrates the S3DIS segmentation results visualization employing our LSNet. The LSNet effectively classifies the majority of points. However, there are notable discrepancies in detail, particularly in differentiating between walls and tables, which aligns with the findings of previous methods (e.g., NeiEA-Net \citep{neiea}).

\subsubsection{Evaluation on Toronto3D}
As summarized in Table \ref{tab:three}, the incorporation of the PAE module yielded a sustained growth pattern in comparison to the baselines. The segmentation performance of the four categories of natural elements, build, pole, and fence has been consistently enhanced. %
This is likely due to the fact that these categories have more significant height structural information compared to ground labels, and the PAE module pushes the allocation of vertical neighboring points. Furthermore, our LSNet exceeded the performance of recent work in OA ($97.5\%$) and mIoU ($83.4\%$). Figure \ref{fig:five} illustrates the results of the segmentation process. The visualization demonstrates the efficacy of our approach, particularly in the recognition of buildings and natural trees. This is attributed to the larger receptive field of our network, which is effective for identifying large objects. Additionally, the introduction of 2D-KNN has enhanced the sensitivity of LSNet to height differences between objects and the ground.

\begin{table*}
	\begin{center}
		\resizebox{\textwidth}{!}{
			\begin{tabular}{@{}lccccccccccccccc@{}}
				\hline
				Method                        & \rotatebox{45}{OA ($\% $)}       & \rotatebox{45}{mIoU ($\% $)}     & \rotatebox{45}{IoUs ($\% $)}     &                         &                                  &                         &                                 &                                  &                         &                                  &                         &                          &                                  &                         &                         \\ \cline{4-16}

				                              &                                  &                                  & \rotatebox{45}{ground}           & \rotatebox{45}{veg}     & \rotatebox{45}{building}         & \rotatebox{45}{wall}    & \rotatebox{45}{bridge}          & \rotatebox{45}{parking}          & \rotatebox{45}{rail}    & \rotatebox{45}{traffic}          & \rotatebox{45}{street}  & \rotatebox{45}{car}      & \rotatebox{45}{footpath}         & \rotatebox{45}{bike}    & \rotatebox{45}{water}   \\
				\hline
				PointNet \citep{pointnet}     & 80.7                             & 23.7                             & 67.9                             & 89.5                    & 80.0                             & 0.0                     & 0.0                             & 3.9                              & 0.0                     & 31.5                             & 0.0                     & 35.1                     & 0.0                              & 0.0                     & 0.0                     \\

				PointNet++ \citep{pointnet++} & 84.3                             & 32.9                             & 72.4                             & 94.2                    & 84.7                             & 2.7                     & 2.0                             & 25.7                             & 0.0                     & 31.5                             & 11.4                    & 38.8                     & 7.1                              & 0.0                     & 56.9                    \\

				SPGraph \citep{spg}           & 76.9                             & 37.2                             & 69.9                             & 94.5                    & 88.8                             & 32.8                    & 12.5                            & 15.7                             & 15.4                    & 30.6                             & 22.9                    & 56.4                     & 0.5                              & 0.0                     & 44.2                    \\

				SparseConv \citep{sparseconv} & 85.2                             & 42.6                             & 74.1                             & 97.9                    & 94.2                             & 63.3                    & 7.5                             & 24.2                             & 0.0                     & 30.1                             & 34.0                    & 74.4                     & 0.0                              & 0.0                     & 54.8                    \\

				KPConv \citep{kpconv}         & 93.2                             & 57.5                             & 87.1                             & \textbf{98.9}           & 95.3                             & \textbf{74.4}           & 28.6                            & 41.3                             & 0.0                     & 55.9                             & \textbf{54.4}           & \textbf{85.6}            & 40.3                             & 0.0                     & \textbf{86.3}           \\

				BAAF-Net \citep{baaf}         & 91.7                             & 59.6                             & 80.0                             & 93.9                    & 94.0                             & 65.7                    & 25.4                            & 63.0                             & 50.2                    & 61.9                             & 42.4                    & 79.7                     & 42.1                             & 0.0                     & 79.6                    \\

				CBL \citep{cbl}               & 91.1                             & 56.2                             & 80.0                             & 96.6                    & 95.6                             & 59.7                    & 20.0                            & 45.4                             & 52.9                    & 62.5                             & 44.9                    & 81.8                     & 19.3                             & 0.0                     & 71.4                    \\

				SCF-Net \citep{scf}           & 91.7                             & 61.3                             & 78.3                             & 90.9                    & 92.4                             & 64.3                    & 35.2                            & 57.0                             & 47.5                    & 63.5                             & 44.5                    & 78.4                     & 44.2                             & \textbf{16.2}           & 83.9                    \\

				MVP-Net \citep{mvpnet}        & 93.3                             & 59.4                             & 85.1                             & 98.5                    & 95.9                             & 66.6                    & 57.5                            & 52.7                             & 0.0                     & 61.9                             & 49.7                    & 81.8                     & 43.9                             & 0.0                     & 78.2                    \\

				LACV-Net \citep{lacv}         & 93.2                             & 61.3                             & 85.5                             & 98.4                    & 95.6                             & 61.9                    & 58.6                            & 64.0                             & 28.5                    & 62.8                             & 45.4                    & 81.9                     & 42.4                             & 4.8                     & 67.7                    \\

				\hline
				RandLA-Net \citep{randla}     & 89.7                             & 52.6                             & 80.1                             & 98.0                    & 91.5                             & 48.8                    & 40.7                            & 51.6                             & 0.0                     & 56.6                             & 33.2                    & 80.1                     & 32.6                             & 0.0                     & 71.3                    \\
				\textbf{+ PAE}                & \textcolor{blue}{93.0}           & \textcolor{blue}{63.6}           & \textcolor{blue}{86.1}           & \textcolor{blue}{98.6 } & \textcolor{blue}{94.8}           & \textcolor{blue}{66.6}  & \textcolor{blue}{60.8}          & \textcolor{blue}{57.6  }         & \textcolor{blue}{29.6}  & \textcolor{blue}{60.4 }          & \textcolor{blue}{48.0 } & \textcolor{blue}{83.4}   & \textcolor{blue}{41.8 }          & \textcolor{blue}{6.2 }  & \textcolor{blue}{80.3 } \\

				\hline
				NeiEA-Net \citep{neiea}       & 91.7                             & 57.0                             & 83.3                             & 98.1                    & 93.4                             & 50.1                    & 61.3                            & 57.8                             & 0.0                     & 60.0                             & 41.6                    & 82.4                     & 42.1                             & 0.0                     & 71.0                    \\
				\textbf{+ PAE}                & \textcolor{blue}{93.7}           & \textcolor{blue}{64.4}           & \textcolor{blue}{\textbf{87.2}}  & \textcolor{blue}{98.7 } & \textcolor{blue}{95.1}           & \textcolor{blue}{72.3}  & \textcolor{blue}{\textbf{63.7}} & \textcolor{blue}{59.0}           & \textcolor{blue}{27.8}  & \textcolor{blue}{ 62.6}          & \textcolor{blue}{51.6}  & \textcolor{blue}{83.0}   & \textcolor{blue}{46.8}           & \textcolor{blue}{4.6}   & \textcolor{blue}{85.2}  \\

				\hline

				LSNet                         & \textcolor{black}{\textbf{94.3}} & \textcolor{black}{\textbf{66.2}} & \textcolor{black}{\textbf{87.2}} & \textcolor{black}{98.6} & \textcolor{black}{\textbf{95.4}} & \textcolor{black}{65.6} & \textcolor{black}{58.6}         & \textcolor{black}{\textbf{64.2}} & \textcolor{black}{50.3} & \textcolor{black}{\textbf{63.8}} & \textcolor{black}{51.0} & \textcolor{black}{84.9 } & \textcolor{black}{\textbf{47.2}} & \textcolor{black}{11.2} & 83.0                    \\
				\hline
			\end{tabular}
		}
	\end{center}
	\caption{SensatUrban semantic segmentation results. Blue font color indicates improved results by adding our PAE module to baseline networks. The best result is marked in bold.}
	\label{tab:four}
\end{table*}

\begin{figure*}
	\begin{center}
		\includegraphics[width=0.8\textwidth]{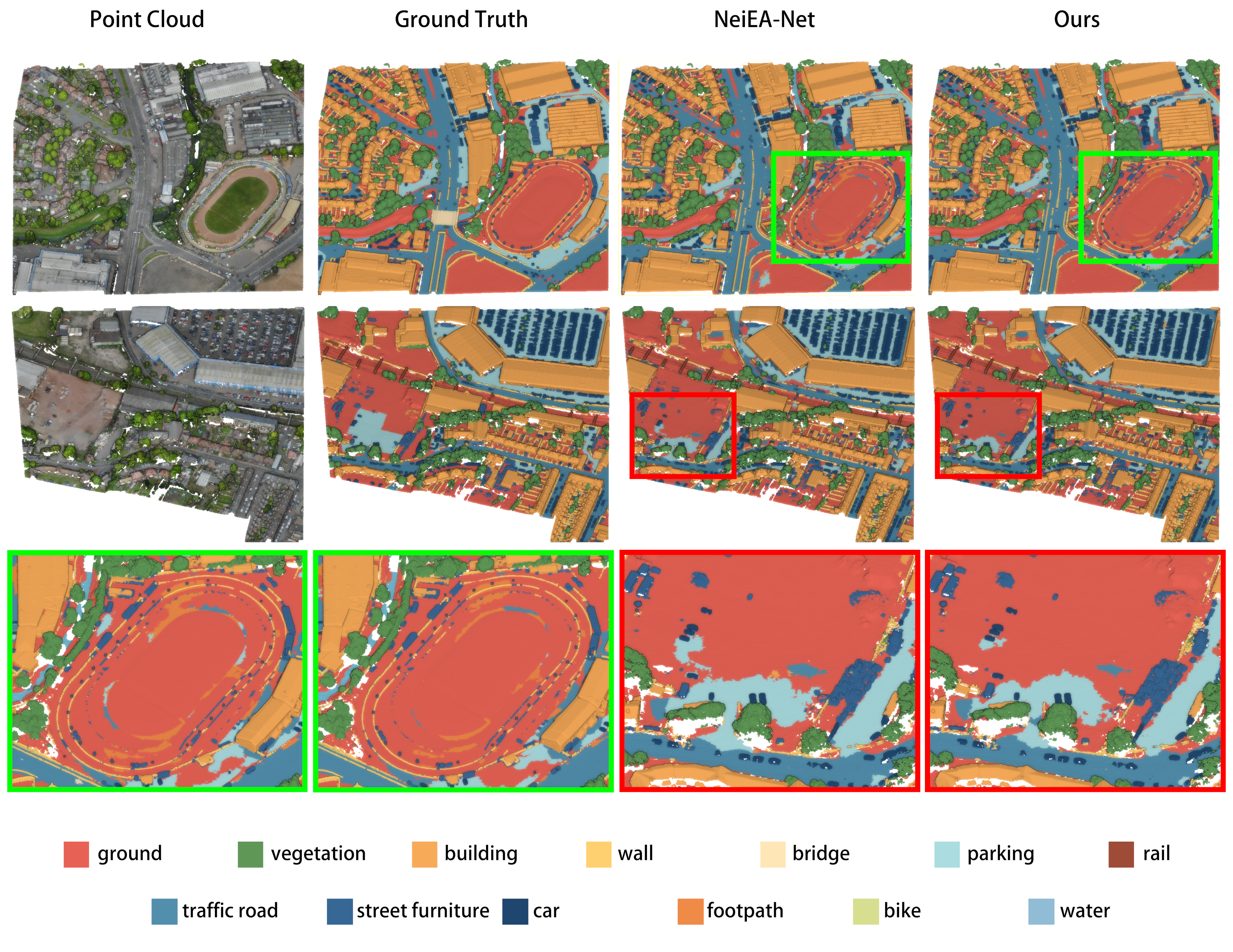}
	\end{center}
	\caption{Visualization of semantic segmentation result of SensatUrban. The rectangles indicate regions where our method achieved superior results compared to others. Third row represents magnified view of corresponding rectangles.}
	\label{fig:sensaturban}
\end{figure*}

\subsubsection{Evaluation on SensatUrban}
Table \ref{tab:four} demonstrated that our PAE module resulted in a notable enhancement of the baseline networks, with an increase of up to 11\% in mIoU. Furthermore, the segmentation of all categories was enhanced, thereby demonstrating the distinctive capabilities of our PAE module on very large-scale urban point clouds. Furthermore, the proposed LSNet achieved a new state-of-the-art performance of 66.2\% mIoU on this extremely large and challenging dataset. As demonstrated in Figure ~\ref{fig:sensaturban}, we display the segmentation outcomes of our approach on the verification set of SensatUrban to better examine its semantic segmentation impact. As observed in this figure, our network effectively recognizes large objects such as ground, buildings, parking, and traffic roads. This is due to LSNet's larger receptive field and the introduction of 2D-KNN, which enhances the network's perception capabilities. This allows for efficient segmentation of planar objects like ground, as well as objects with significant height differences from the plane, such as buildings. 

\subsection{Ablation Study}
We conducted ablation experiments on various modules of LSNet to better demonstrate its effectiveness. All the experiments below are conducted on the SensatUrban dataset.

\subsubsection{Ablation Study of PAE Module}
The objective is to examine the influence of diverse neighborhood search combinations on local feature learning, while maintaining the integrity of the LSAP block. The following ablation experiments are conducted: (1 - 4) utilize a single neighborhood search algorithm for local feature learning; (5 - 6) employ two distinct neighborhood search algorithms; (7) utilizes three different neighborhood search algorithms to aggregate local feature learning; (8) local feature learning of two neighborhood search algorithms used by our PAE module.

\begin{table}[]
	\begin{center}

		\tabcolsep=1.5mm
		\begin{tabular}{c|ccc|c|c}
			\hline
			Model & \multicolumn{3}{c|}{2D-KNN}                    & \multicolumn{1}{l|}{3D-KNN}                    & mIoU                                                                                      \\
			      & $\left[ {y,z} \right]$ & $\left[ {x,z} \right]$ & $\left[ {x,y} \right]$ & $\left[ {x,y,z} \right]$ & ($\% $)       \\
			\hline
			1     & \checkmark                                     &                                                &                                                &                          & 51.9          \\
			2     &                                                & \checkmark                                     &                                                &                          & 52.4          \\
			3     &                                                &                                                & \checkmark                                     &                          & 54.5          \\
			4     &                                                &                                                &                                                & \checkmark               & 61.2          \\ \hline
			5     & \checkmark                                     &                                                &                                                & \checkmark               & 60.8          \\
			6     &                                                & \checkmark                                     &                                                & \checkmark               & 60.5          \\ \hline
			7     & \checkmark                                     & \checkmark                                     & \checkmark                                     & \checkmark               & 58.2          \\
			\hline
			8     &                                                &                                                & \checkmark                                     & \checkmark               & \textbf{66.2} \\ \hline
		\end{tabular}
	\end{center}
	\caption{Ablation study of local feature learning using different neighborhood search algorithms in PAE module.}
	\label{tab:five}
\end{table}

As demonstrated in Table \ref{tab:five}, the application of a single neighborhood search algorithm for local feature learning results in a constrained semantic segmentation performance. In particular, the application of the 3D-KNN algorithm for local feature learning resulted in an improvement of 9.3\% in mIoU compared to the use of the 2D-KNN algorithm (in comparison to YZ). This is due to the fact that employing solely the 2D-KNN algorithm results in a considerable reduction of the structural features inherent to the point cloud. The combination of 3D-KNN with YZ or XZ 2D-KNN was detrimental to the model's mIoU. However, when combining 3D-KNN with XY 2D-KNN, the segmentation performance of the model reached its optimal level, with mIoU (+5\% compared to 3D-KNN only). This is due to the fact that, in practical situations, the objects usually have an elongated shape along the vertical dimension in urban environments. This experimental outcome serves to illustrate the efficacy of the PAE module.

\begin{table}[]
	\begin{center}
		\resizebox{\linewidth}{!}{
			\begin{tabular}{clccc}
				\hline
				Model & Per-attention & Per-batch time & OA           & mIoU          \\
				      &               & (ms)           & ($\% $)       & ($\% $)       \\ \hline
				1     & k = 9         & 367            & 91.9          & 58.9          \\
				2     & k = 16        & 532            & 92.7          & 62.6          \\
				3     & k = 25        & 673            & 93.4          & 65.6          \\
				5     & k = 36        & 827            & 93.5          & 64.5          \\ \hline
				4     & k = 25 + LSAP & 506            & \textbf{94.3} & \textbf{66.2} \\
				6     & k = 36 + LSAP & 546            & 94.0          & 66.1          \\

				\hline
			\end{tabular}
		}
	\end{center}

	\caption{Ablation study of LSAP with different neighboring sizes (i.e., K value in KNN).}
	\label{tab:ff}
\end{table}

\subsubsection{Ablation Study of LSAP Block}
The number of neighbors utilized by the KNN algorithm, which is employed by the majority of point cloud deep learning networks, directly influences the network's capacity to perceive local contextual information. In this study, we conducted ablation experiments to examine the impact of varying the K value in KNN for local feature learning. Subsequently, the LASP block, which is specifically designed for large receptive fields, was incorporated into the tested 25 and 36 neighboring point situations.

As illustrated in Table \ref{tab:ff}, it is evident that as the receptive field increases, both the segmentation performance and the running time of the model demonstrate a corresponding enhancement. In comparison to local feature learning with k=9, the model demonstrated an improvement in mIoU by 5.6\%, yet the per-batch time exhibited a more than twofold increase at k=36. However, at k=25, mIoU exhibited a 1.1\% increase relative to k=36, indicating that an overly large receptive field may potentially lead to an increase in redundant information and a tendency towards overfitting, thereby affecting the segmentation performance of the model.

For k=25 and k=36, respectively, the proposed method exhibited superior performance in terms of mIoU (+0.6\%) and per-batch time (-167 ms) in comparison to k=25 without LSAP, and mIoU (+1.6\%) and per-batch time (-271 ms) in comparison to k=36 without LSAP. These results illustrate that our approach not only enhances the segmentation capabilities of the model but also alleviates computational constraints, leading to a notable increase in processing speed of approximately 38.8\%.

\begin{table}[]
	\begin{center}
		\begin{tabular}{cccc}
			\hline
			Model & Method                      & OA (\%)       & mIoU (\%)     \\ \hline
			1     & Replace FMA with MLP        & 92.7          & 62.6          \\
			2     & Mean pooling                & 93.9          & 65.7          \\

			\hline
			3     & \textbf{Max pooling (Ours)} & \textbf{94.0} & \textbf{66.2} \\ \hline
		\end{tabular}

	\end{center}

	\caption{Ablation study of FMA module.}
	\label{tab:seven}
\end{table}

\subsubsection{Ablation Study of FMA Module}
In the decoding layers, the features were upsampled using the KNN algorithm. However, this method may result in the loss of some effective features. To address this issue, we conducted the following ablation experiments: (1) the decoding layer of the network was modified to only use MLP; (2) an average aggregation operation was added; (3) a max aggregation operation was added.

As demonstrated in Table \ref{tab:seven}, the exclusive utilization of MLP resulted in a decline in network performance, exhibiting a -3.6\% reduction in mIoU compared to the proposed approach. The transmission of low-resolution features is not as effective during upsampling. To enhance this aspect, we evaluated the semantic segmentation performance of the network when incorporating distinct pooling layers into FMA. The findings indicate that the max pooling layer exhibited the most notable improvement in network performance, with mIoU (+0.5\% compared to mean pooling).

\section{Discussion}

\begin{figure*}[htbp]
	\begin{minipage}[t]{0.5\linewidth}
		\centering
		\includegraphics[width=\textwidth]{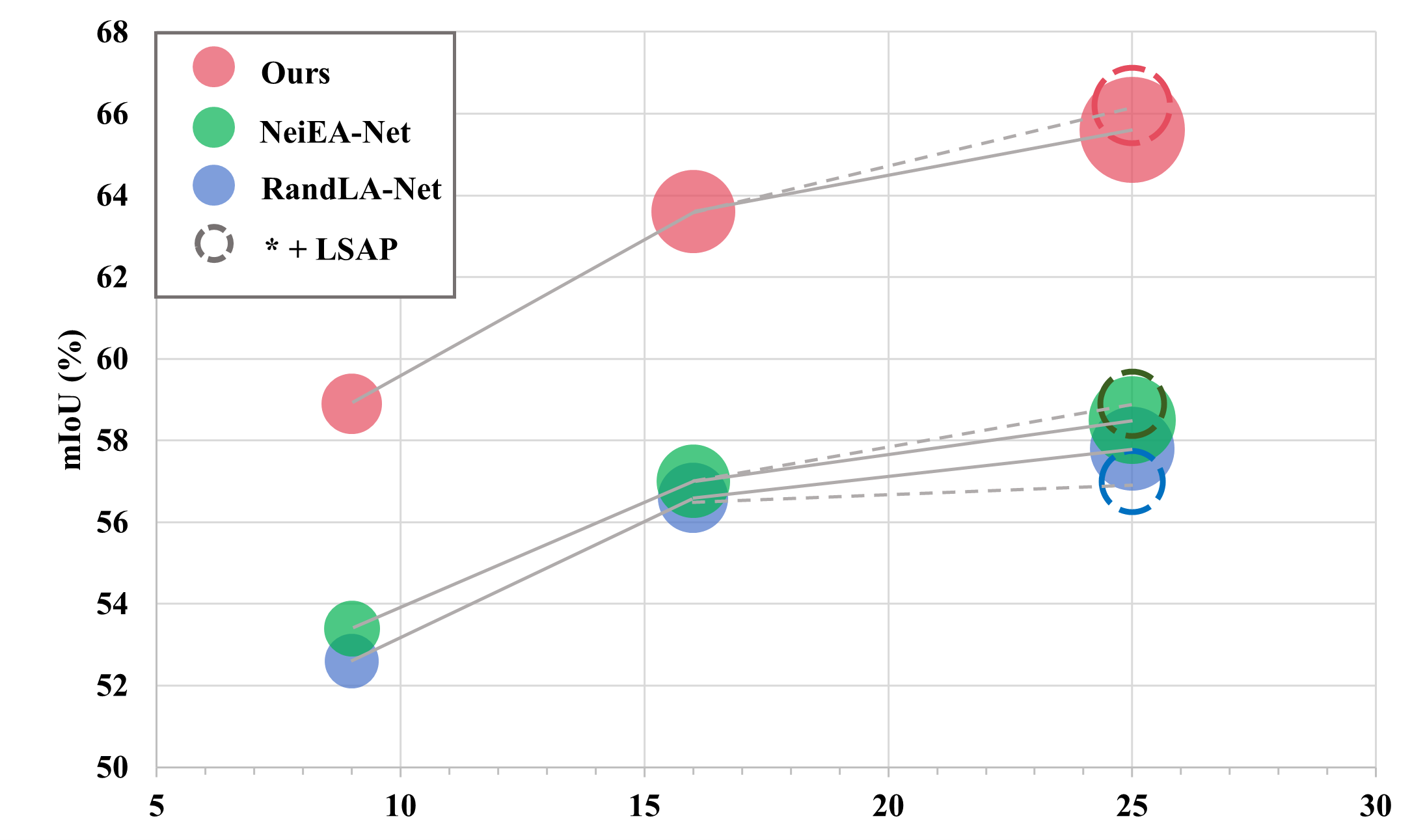}
	\end{minipage}%
	\begin{minipage}[t]{0.5\linewidth}
		\centering
		\includegraphics[width=\textwidth]{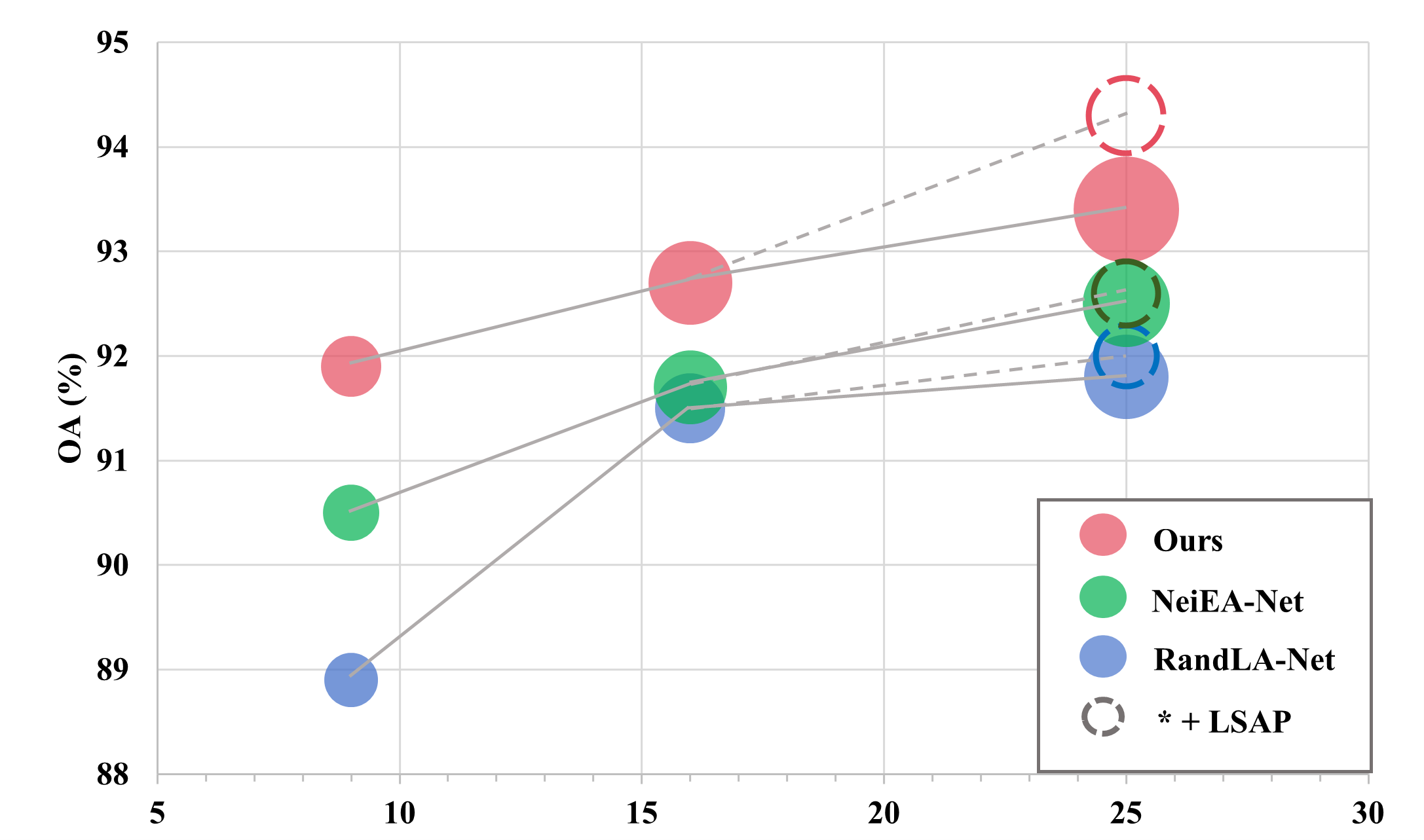}
	\end{minipage}
	\caption{Impacts of receptive field (i.e., K value in X axis) on model performance and efficiency. Circle size denotes computational time. Dashed circle indicated model with the LSAP mechanism.}
	\label{fig:both_images}
\end{figure*}

\subsection{Revisite the Role of Receptive Field}
\label{role}

In point cloud processing, the receptive field is the parameter that determines the extent of the space surrounding a given point that is taken into account when computing its feature representation. In the context of large-scale analysis, a larger receptive field is typically preferred to capture global structures and contextual information \citep{randla}. However, the use of a large receptive field may also have adverse effects, including an increased computational load, the loss of fine-grained details, and an overfitting to the training data, which may result in a lack of generalization to new point clouds. This suggests that a balance should be sought between these two opposing factors.

To gain further insight into this issue, we conducted an investigation into the impact of receptive field on existing networks and our own network. As demonstrated in Figure \ref{fig:both_images}, both RandLA-Net \citep{randla} and NeiEA-Net \citep{neiea} exhibited enhanced performance with an expanded neighborhood size, ranging from 9 to 25. This observation was also evident in our LSNet model that without the LSAP mechanism. However, while the mIoU could be increased by over 5\%, the computational time increased by 55\% for RandLA-Net and NeiEA-Net, and 74\% for our model. Nevertheless, the incorporation of the LSAP mechanism resulted in a notable reduction in computational time for all models, accompanied by an increase in mIoU, with the exception of RandLA-Net. The computational time for K=25 was observed to be lower than that of K=16 with the integration of the LSAP mechanism. This observation provides compelling evidence that our method can effectively expand the model's receptive field, thereby enhancing its ability to learn features while maintaining a low computational burden and avoiding overfitting.

\subsection{Towards Urban-scale Point Cloud Analysis}
Despite the considerable advancements made in point cloud analysis in recent years, particularly with the advent of deep learning techniques powered by artificial intelligence, it is important to note that the majority of existing research has focused on small-scale objects or scenes, with the primary objective being the development of generic point cloud deep learning networks. With the advancement of modern point cloud acquisition techniques, such as mobile laser scanning and photogrammetry, the use of these techniques in urban-scale outdoor applications has grown rapidly. This has led to a need for practical solutions that can automatically analyze extremely large-scale point clouds from extremely complex outdoor scenes.  

In light of \cite{randla}, numerous studies have been conducted in order to address this issue. These studies focused on two key areas: computational burden and local contextual learning. A key observation of urban-scale scenes is that the dominant objects, such as buildings and trees, are large in vertical dimension and mostly distributed along the horizontal plane (Figure \ref{sensatori}). This generally requires a large receptive field in the model, which may potentially give rise to the issues discussed in section \ref{role}. In this study, our designed PAE module incorporates an additional 2D-KNN channel, which effectively compresses the objects along the vertical dimension, thereby allowing for the allocation of a greater number of neighboring points from the vertical dimension. 

Our work represents a significant advancement in large-scale urban point cloud analysis, as evidenced by the development of novel strategies tailored to this domain. The state-of-the-art results on the Toronto3D and SensatUrban datasets demonstrate that the strategies developed by our method are particularly effective for large-scale urban data. It is also noteworthy that this enhancement does not compromise computational efficiency, which is essential for large-scale data processing.

\subsection{Limitations and Future Work}
While our method yielded promising outcomes on the experimental datasets, particularly the exceptionally large SensatUrban dataset, it is important to acknowledge that there is still room for enhancement. Firstly, although the receptive field was efficiently and effectively expanded by our method, it should be noted that the receptive field is fixed and therefore might not be optimal for all objects. In particular, in complex urban environments, objects exhibit a range of sizes, which suggests that the optimal receptive field may also vary. It would be optimal to learn an adaptive receptive field for different objects \citep{wei2017learning}. Additionally, the proposed approach employs a fully supervised methodology. The cost of data labeling is considerable in large scenes, suggesting an avenue for further research into the application of weakly supervised or self-supervised approaches to handle sparse point cloud data. The combination of these approaches with modern machine learning algorithms has the potential to mitigate data dependency.

\section{Conclusion}
This paper introduces the local split attention pooling mechanism and a parallel feature aggregate module, which facilitate the obtaining of a large receptive field in point cloud deep learning networks. The objective is to enhance the model's capacity to capture comprehensive contextual data while ensuring computational feasibility and preventing model overfitting. In light of these novel mechanisms and module designs, we put forth a novel network, designated LSNet, for large-scale point cloud semantic segmentation. Experiments were conducted on three large-scale datasets, and the results demonstrate that our model exhibits superior performance. Moreover, the experiments demonstrate that the PAE module is highly portable and can be readily transferred to existing frameworks to enhance model performance. Ablation studies were also conducted to verify the effectiveness of the designed model components.  Our work contributes to the effective expansion of the model's receptive field to accommodate the complexities of large point clouds. Future work will focus on enabling the model to retain fine-grained features while maintaining a large receptive field.

\printcredits
\section*{Declaration of competing interest}
The authors declare that they have no known competing financial interests or personal relationships that could have appeared to influence the work reported in this paper.
\section*{Data availability}
The code for our key contributions is availale at: will be added if accepted.%
\section*{Acknowledgments}
This work is supported by the National Natural Science Foundation of China (No. 42101330), the National Key Research and Development Program of China (No. 2021YFF0704600), and the Key Research and Development Program of Shaanxi Province (No. 2023-YBSF-452).

\bibliographystyle{model1-num-names}
\bibliography{ref}

\end{document}